%% file: conference_101719.tex
\documentclass[conference]{IEEEtran}
\IEEEoverridecommandlockouts
\usepackage{cite}
\usepackage{amsmath,amssymb,amsfonts}
\usepackage{algorithmic}
\usepackage{graphicx}
\usepackage{textcomp}
\def\BibTeX{{\rm B\kern-.05em{\sc i\kern-.025em b}\kern-.08em
    T\kern-.1667em\lower.7ex\hbox{E}\kern-.125emX}}

\usepackage{booktabs}
\usepackage{multirow}
\usepackage[table]{xcolor}
\usepackage{float}
\usepackage{url}
\usepackage{subcaption}
    
\begin{document}

\title{CGRL: Concept-Guided Pruning and Representation Learning for Whole-Slide Image Classification}

% Anonymous version for double-blind review
% \author{
% \IEEEauthorblockN{
% Thuc Huynh, Tuan Le, Doanh C. Bui\textsuperscript{*}
% }
% \IEEEauthorblockA{
% \textit{University of Information Technology, VNU-HCM} \\
% Ho Chi Minh City, Vietnam \\
% 23521555@gm.uit.edu.vn, tuanlv@uit.edu.vn, doanhbc@uit.edu.vn
% }

% \thanks{\textsuperscript{*} Corresponding author: Doanh C. Bui, doanhbc@uit.edu.vn.}
% }

% \maketitle

\author{
\IEEEauthorblockN{
Thuc Huynh\textsuperscript{1,2},
Tuan Le\textsuperscript{1,2}, and
Doanh C. Bui\textsuperscript{1,2,\ensuremath{\dagger}}
}

\IEEEauthorblockA{
\textsuperscript{1}\textit{
University of Information Technology, Ho Chi Minh City, Vietnam
}
}

\IEEEauthorblockA{
\textsuperscript{2}\textit{
Vietnam National University Ho Chi Minh City, Ho Chi Minh City, Vietnam
}
}

\IEEEauthorblockA{
23521555@gm.uit.edu.vn,
\{tuanlv, doanhbc\}@uit.edu.vn
}

\IEEEauthorblockA{
\textsuperscript{\ensuremath{\dagger}}
Corresponding author: doanhbc@uit.edu.vn.
}
}

\maketitle

\input{sections/01_abstract}

\input{sections/02_introduction}

\input{sections/03_related_work}
\input{sections/04_method}
\input{sections/05_experiments}

\input{sections/06_discussion}
\input{sections/07_conclusion}
\section*{Acknowledgment}
This research was supported by The VNUHCM-University of Information Technology's Scientific Research Support Fund.
\bibliographystyle{IEEEtran}
\bibliography{refs/references}
\end{document}

%% file: sections/01_abstract.tex
\begin{abstract}
Weakly supervised whole-slide image (WSI) classification has become a major paradigm in computational pathology, since slide-level labels are substantially easier to obtain than dense region annotations. 
Most existing multiple instance learning (MIL) methods aggregate large bags of patch embeddings based primarily on visual cues, which often leads to two limitations: redundancy caused by numerous non-informative patches and weak alignment between instance features and class-level disease semantics. 
To address these issues, we propose \emph{Concept-Guided Pruning and Representation Learning} (CGRL), a simple and effective framework for weakly supervised WSI classification. 
CGRL introduces class-level concept prototypes derived from disease prompts and incorporates them into the MIL pipeline in two complementary ways. 
First, a \emph{concept-relevance pruning} module ranks patch instances according to their similarity to class-level concepts and retains the top-$K$ concept-relevant patches for downstream MIL aggregation. 
Second, a \emph{concept-guided contrastive representation learning} module constructs class-wise positive and negative patch sets from the same similarity matrix and optimizes them with target-class, symmetric auxiliary, and cross-class separation objectives, thereby regularizing the semantic structure of the projected concept space. 
We evaluate CGRL on two benchmark TCGA datasets, TCGA-BRCA and TCGA-NSCLC, using multiple representative MIL methods. 
Experimental results show that CGRL improves performance in several model–dataset combinations, with gains that depend on the downstream MIL model and dataset, with especially clear gains in accuracy and macro-F1, while also reducing the computational burden through concept-relevance pruning. 
These findings demonstrate that class-level semantic concepts provide an effective and practical prior for both patch selection and representation learning in weakly supervised computational pathology. Implementation is available at \url{https://github.com/ThucHuynh44/CGRL}.
\end{abstract}

\begin{IEEEkeywords}
whole-slide image analysis, weakly supervised learning, multiple instance learning, representation learning, cancer subtyping
\end{IEEEkeywords}

%% file: sections/02_introduction.tex
\section{Introduction}

% \begin{figure}[t]
%     \centering
%     \includegraphics[width=\linewidth,page=1]{figures/flops.pdf}
%     \caption{FLOPs for representative MIL methods on the TCGA-NSCLC dataset}
%     \label{fig:flops}
% \end{figure}

Histopathological examination of whole-slide images (WSIs) plays a central role in cancer diagnosis and subtype identification. 
With the rapid digitization of pathology workflows, computational analysis of WSIs has become increasingly important for improving diagnostic efficiency and reproducibility~\cite{lu2021data,gadermayr2024milreview}. 
However, a WSI is typically of gigapixel scale and may contain tens of thousands of image patches, making direct end-to-end supervised learning extremely challenging, while detailed region-level annotation is costly and time-consuming~\cite{lu2021data,shao2021transmil,gadermayr2024milreview}. 
This has motivated growing interest in weakly supervised learning methods that rely only on slide-level labels~\cite{lu2021data,gadermayr2024milreview}.

\begin{figure*}[t]
    \centering
    % Hình a (bên trái)
    \begin{subfigure}[b]{0.48\linewidth}
        \centering
        \includegraphics[width=\linewidth,page=1]{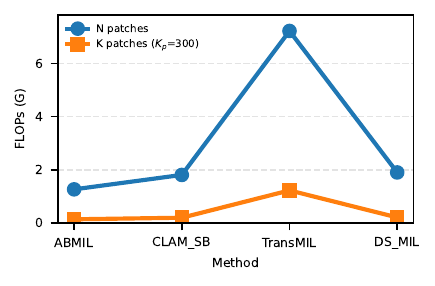}
        \caption{FLOPs for representative MIL methods on the TCGA-NSCLC dataset}
        \label{fig:flops}
    \end{subfigure}
    \hfill % Tạo khoảng trống ngang tự động giữa 2 hình
    % Hình b (bên phải)
    \begin{subfigure}[b]{0.48\linewidth}
        \centering
        \includegraphics[width=\linewidth,page=1]{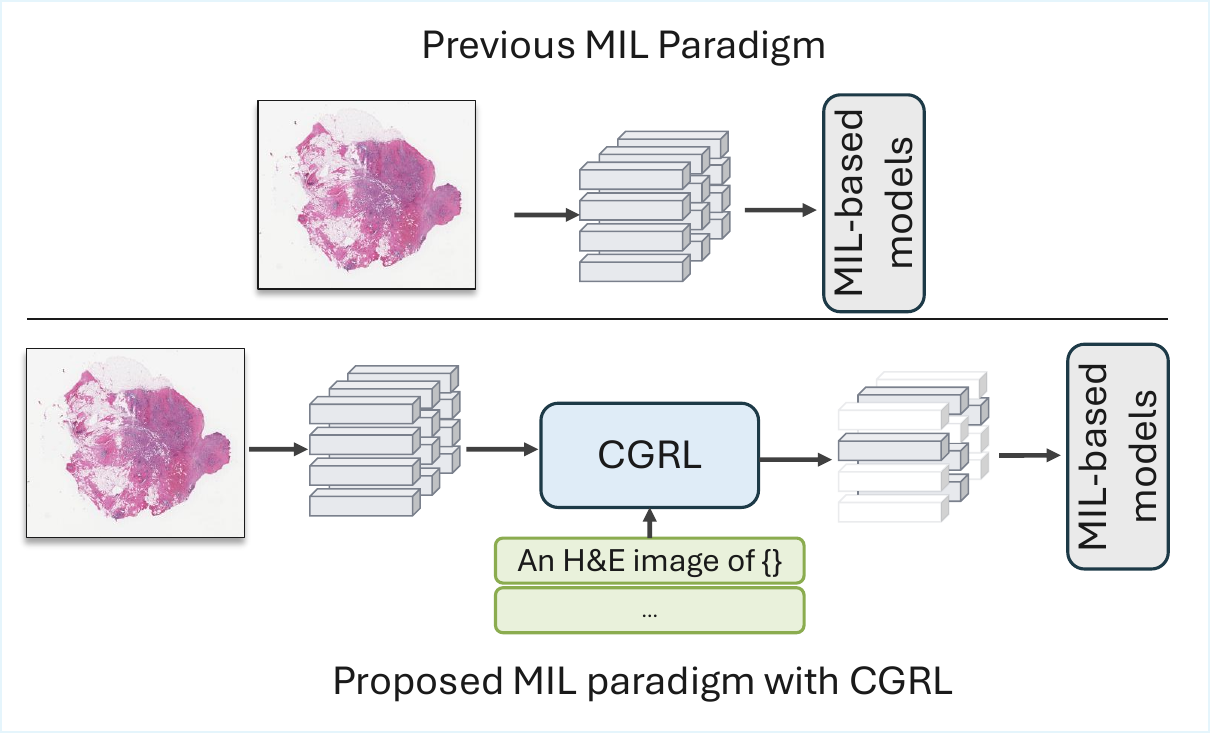}
        \caption{Comparison between the previous MIL paradigm and the proposed MIL framework with CGRL}
        \label{fig:mil_paradigm}
    \end{subfigure}
    
    \caption{Overall comparison of MIL methods and paradigms.}
    \label{fig:combined_mil}
\end{figure*}

Multiple instance learning (MIL) has become the dominant paradigm for weakly supervised WSI classification~\cite{ilse2018attention,lu2021data,li2021dual,shao2021transmil,gadermayr2024milreview}. 
In MIL-based methods, a slide is represented as a bag of patch embeddings, and a bag-level predictor aggregates instance features to produce the final diagnosis. 
Despite their success, most existing MIL approaches still rely primarily on visual representations learned from image patches, which leads to two limitations. 
First, the original WSI bag usually contains a large number of redundant or weakly informative patches, which increases computational cost and may dilute diagnostically relevant signals (see Fig.~\ref{fig:flops})~\cite{gadermayr2024milreview,liu2024pamil,zhao2025ptcmil}. 
Second, in the absence of explicit semantic guidance, the learned instance space may not align well with high-level disease concepts, especially when the distinction between subtypes depends on subtle morphological cues~\cite{gadermayr2024milreview,liu2024pamil,zhao2025ptcmil}.

Recent advances in vision--language learning provide a promising direction for addressing these limitations~\cite{radford2021clip,lu2024conch,ding2025multimodal}. 
Pretrained image and text encoders enable visual regions and textual descriptions to be mapped into a shared semantic space, allowing disease-specific prompts to serve as compact semantic anchors for pathology analysis~\cite{lu2024conch,ding2025multimodal,huang2024free}. 
However, effectively incorporating such semantic concepts into weakly supervised MIL remains nontrivial, since an effective framework should support both informative patch selection and semantic regularization during training.

To this end, we propose \emph{Concept-Guided Pruning and Representation Learning} (CGRL), a simple and effective framework for weakly supervised WSI classification. As illustrated in Fig.~\ref{fig:mil_paradigm}, CGRL augments the conventional MIL pipeline with a concept-guided module that supports both concept-relevance pruning and concept-guided representation learning.
Given a bag of patch embeddings, CGRL introduces class-level concept prototypes derived from disease prompts and integrates them into the MIL pipeline in two complementary ways. 
First, CGRL performs \emph{concept-relevance pruning} to rank patches according to their similarity to class-level concepts and retain only the top-$K$ concept-relevant instances for downstream MIL aggregation, thereby reducing redundancy and improving computational efficiency. 
Second, CGRL performs \emph{concept-guided representation learning} by constructing class-wise positive and negative patch sets from the concept similarity matrix and optimizing them with contrastive objectives. 
Thus, class-level semantic concepts act both as selection cues and as supervisory signals that regularize the projected concept space. During inference, only concept-relevance pruning and MIL-based slide classification are retained, making CGRL a lightweight add-on for existing MIL models.

We evaluate CGRL on two benchmark TCGA datasets, TCGA-BRCA~\cite{tcga2012brca} and TCGA-NSCLC, using several representative MIL methods. 
Experimental results show that CGRL improves multiple model--dataset combinations, with particularly clear gains in threshold-dependent metrics such as accuracy and macro-F1 in several settings. 
In addition, by pruning the input bag before MIL aggregation, CGRL reduces the computational burden compared with using all extracted patches. 
These results suggest that class-level semantic concepts provide a useful prior for both instance selection and representation learning in weakly supervised pathology.

In summary, the main contributions of this work are as follows:
\begin{itemize}
    \item We propose CGRL, a concept-guided framework that injects class-level semantic priors into weakly supervised WSI classification.
    \item We introduce a unified design that couples concept-relevance pruning with concept-guided representation learning, enabling both informative patch selection and improved semantic structure of instance features.
    \item We demonstrate that CGRL can enhance multiple MIL methods across TCGA datasets while also reducing computational cost through concept-relevance pruning.
\end{itemize}

% \begin{figure}[t]
%     \centering
%     \includegraphics[width=\linewidth,page=1]{figures/thumbnail_cropped.pdf}
%     \caption{Comparison between the previous MIL paradigm and the proposed MIL framework with CGRL}
%     \label{fig:mil_paradigm}
% \end{figure}

%% file: sections/03_related_work.tex
\section{Related Work}
Multiple instance learning (MIL) is the dominant framework for weakly supervised WSI classification, where a slide is represented as a bag of patch embeddings and mapped to a slide-level label through bag-level aggregation. Representative methods such as ABMIL, CLAM, DS-MIL, and TransMIL improve bag modeling through attention-based pooling, clustering constraints, dual-stream aggregation, and long-range dependency modeling, respectively~\cite{ilse2018attention,lu2021data,li2021dual,shao2021transmil}. Despite their success, prior studies have shown that MIL for digital pathology still faces several challenges, including redundant instances, weak instance-level semantics, and high computational cost~\cite{gadermayr2024milreview}.

To address these limitations, recent works have explored patch selection, compression, clustering, and prototype-based aggregation before or during MIL inference. For example, PAMIL improves bag modeling with prototype attention, while CICS reduces redundant computation through content-aware compression and selection~\cite{liu2024pamil,zheng2026content}. PTCMIL further enhances bag representation via prompt-token clustering and structured instance modeling~\cite{zhao2025ptcmil}. In parallel, vision--language learning provides a promising way to connect visual pathology patterns with disease semantics through shared image--text representations~\cite{radford2021clip}. Pathology-specific foundation models such as CONCH and TITAN, together with concept-guided prompt adaptation methods such as CATE, have demonstrated the value of semantic priors in pathology analysis~\cite{lu2024conch,ding2025multimodal,huang2024free}. Different from prior work, CGRL uses class-level concept prototypes in a unified manner for both concept-relevance pruning and contrastive representation learning. This design allows semantic concepts to act simultaneously as patch-selection cues and supervisory signals, while remaining a lightweight plug-and-play module built on frozen image and text embeddings.

%% file: sections/04_method.tex
\section{Method}
\begin{figure*}[t]
    \centering
    \includegraphics[width=\linewidth,page=1]{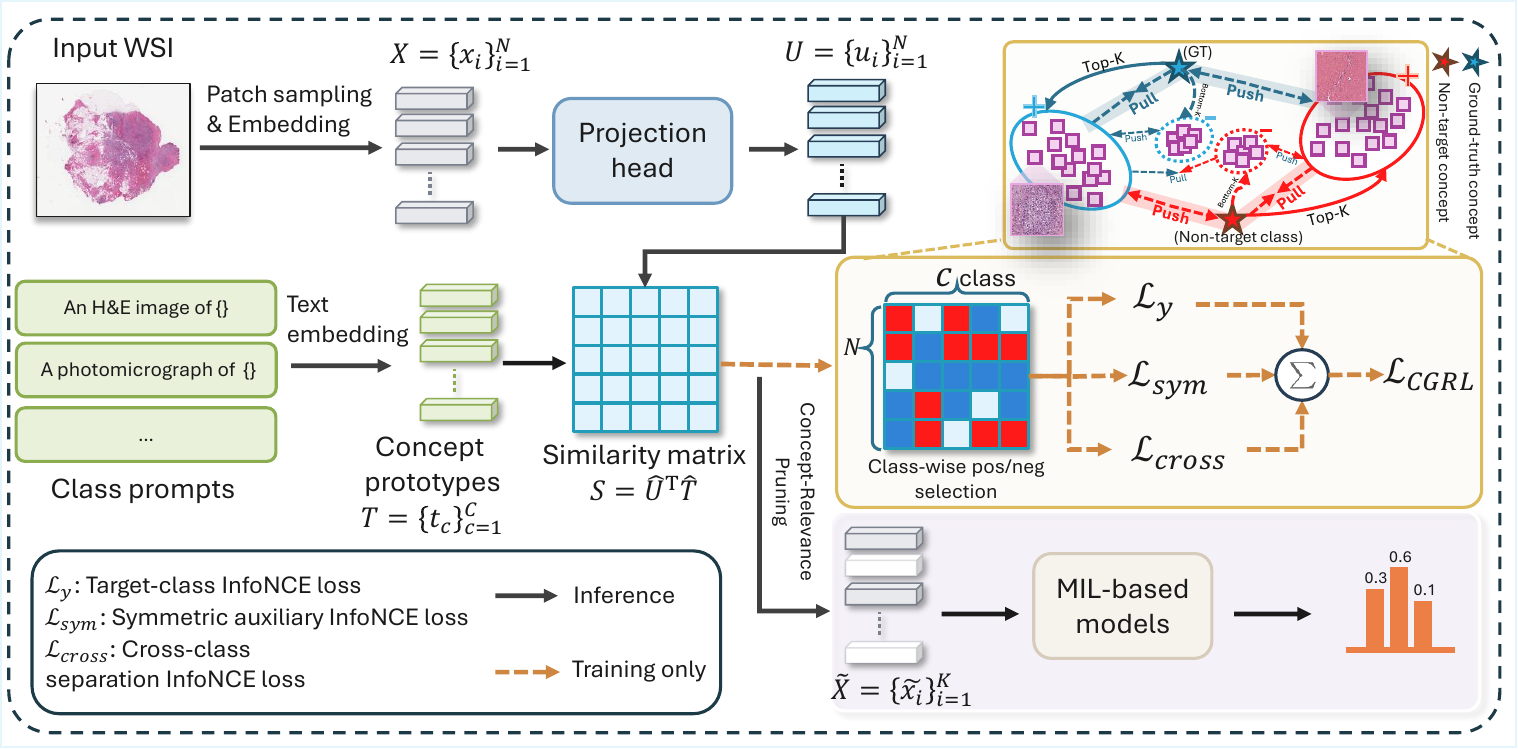}
    \caption{\textbf{Overview of the proposed Concept-Guided Pruning and Representation Learning (CGRL) framework.}
Given a WSI, patch embeddings and class-level concept prototypes are used to compute a patch-to-concept similarity matrix. Based on this similarity, CGRL performs concept-relevance pruning for downstream MIL classification and concept-guided representation learning during training. During inference, only concept-relevance pruning and MIL-based slide classification are retained.}
    \label{fig:framework}
\end{figure*}

\subsection{Method Overview}

We propose \emph{Concept-Guided Pruning and Representation Learning} (CGRL) for weakly supervised whole-slide image classification, as illustrated in Fig.~\ref{fig:framework}. Given a WSI, we first extract a bag of patch embeddings
\begin{equation}
X=\{x_i\}_{i=1}^{N}, \qquad x_i \in \mathbb{R}^{D},
\end{equation}
where $N$ is the number of patches and $D$ is the feature dimension.

We further introduce class-level concept prototypes
\begin{equation}
T=\{t_c\}_{c=1}^{C}, \qquad t_c \in \mathbb{R}^{D_c},
\end{equation}
where each prototype $t_c$ is obtained by encoding class-specific textual prompts and serves as a semantic anchor for class $c$.

A learnable projection head maps patch embeddings into the concept space to produce projected features $U=\{u_i\}_{i=1}^{N}$. Based on $U$ and $T$, we compute a patch-to-concept similarity matrix $S$, which is used for both patch selection and representation learning. First, CGRL performs \emph{concept-relevance pruning} by scoring each patch according to its relevance to the concept prototypes and retaining only the top-$K_p$ most concept-relevant patches. The resulting compact bag is then forwarded to a downstream MIL model for slide-level prediction. Second, CGRL performs \emph{concept-guided representation learning}. Using the same similarity matrix $S$, the model constructs class-wise positive and negative patch sets in the projected concept space and optimizes them with three contrastive objectives: $\mathcal{L}_{y}$, $\mathcal{L}_{sym}$, and $\mathcal{L}_{cross}$. These losses form the overall concept-guided representation learning loss $\mathcal{L}_{\mathrm{CGRL}}$.

During training, CGRL jointly optimizes the MIL classification objective and $\mathcal{L}_{\mathrm{CGRL}}$. During inference, the contrastive learning component is removed, and the model retains only concept-relevance pruning followed by MIL-based slide classification.

\subsection{Concept-relevance pruning and concept-guided representation learning}

\paragraph{Concept-relevance pruning}
To align patch features with concept prototypes, we first map each patch embedding into the concept space using a lightweight learnable projection head $g_{\phi}(\cdot)$:
\begin{equation}
u_i = g_{\phi}(x_i), \qquad u_i \in \mathbb{R}^{D_c}.
\end{equation}
Let $U=\{u_i\}_{i=1}^{N}$ denote the projected patch features. We then compute the similarity matrix between projected patch features and concept prototypes as
\begin{equation}
S = \hat{U}\hat{T}^{\top},
\end{equation}
where $\hat{U}$ and $\hat{T}$ denote the $\ell_2$-normalized versions of $U$ and $T$, respectively. Equivalently, each element of $S$ can be written as
\begin{equation}
S_{i,c} = \frac{u_i^\top t_c}{\|u_i\|_2 \|t_c\|_2},
\end{equation}
which corresponds to the cosine similarity between patch $i$ and concept prototype $t_c$. Based on this similarity matrix, we define the concept relevance score of each patch as the maximum similarity across all concept prototypes:
\begin{equation}
r_i = \max_{c \in \{1,\dots,C\}} S_{i,c}.
\end{equation}
We then retain the top-$K_p$ most concept-relevant patches:
\begin{equation}
\mathcal{I}_{K_p} = \operatorname{TopK}_{K_p}(\{r_i\}_{i=1}^{N}),
\end{equation}
and construct the pruned bag
\begin{equation}
\tilde{X} = \{\tilde{x}_k\}_{k=1}^{K_p}, \qquad \tilde{x}_k = x_{i_k}, \; i_k \in \mathcal{I}_{K_p}.
\end{equation}
Importantly, concept matching is performed in the projected concept space, whereas the original patch embeddings are preserved and forwarded to the downstream MIL model for slide-level prediction. After selection, the retained indices are sorted to preserve the original patch order.

\paragraph{Concept-guided representation learning}
In parallel with concept-relevance pruning, the same similarity matrix $S$ is further used to construct class-wise positive and negative sets for contrastive supervision in the projected concept space. Importantly, these contrastive objectives are computed from the projected patch features $U=\{u_i\}_{i=1}^{N}$ of the original bag, while the pruned bag $\tilde{X}$ is reserved for downstream MIL-based slide classification. For each class $c$, we select the top-ranked and bottom-ranked instances according to the corresponding similarity scores:
\begin{equation}
\mathcal{P}_c = \operatorname{TopK}_{K_{pos}}(S_{:,c}), \qquad
\mathcal{N}_c = \operatorname{BottomK}_{K_{neg}}(S_{:,c}),
\end{equation}
where $\mathcal{P}_c$ and $\mathcal{N}_c$ denote the indices of the most relevant and least relevant projected patch features with respect to concept prototype $t_c$, respectively.

Based on the selected samples, for an input WSI with ground-truth class $y$, we formulate three complementary contrastive objectives. Specifically, $\mathcal{L}_y$ encourages target-relevant projected features to align with the prototype $t_y$ of the ground-truth class; $\mathcal{L}_{sym}$ regularizes the projected feature space by aligning the most concept-relevant projected features of each non-target class $c \neq y$ with their corresponding concept prototypes while separating them from concept-irrelevant ones; and $\mathcal{L}_{cross}$ enhances inter-class discrimination by separating target-relevant projected features from projected features that are highly relevant to competing classes.

We adopt the InfoNCE objective \cite{oord2018representation} for concept-guided contrastive learning. Given a set of queries $Q=\{q_i\}_{i=1}^{N_q}$, a positive key $k^{+}$, and a set of negative keys $\mathcal{M}=\{k^-_m\}_{m=1}^{M}$, the loss is defined as
\begin{multline}
\mathcal{L}_{\mathrm{InfoNCE}}(Q,k^{+},\mathcal{M}) = -\frac{1}{N_q}
\\\sum_{i=1}^{N_q}
\log
\frac{
\exp\left(\operatorname{sim}(q_i,k^{+})/\tau\right)
}{
\exp\left(\operatorname{sim}(q_i,k^{+})/\tau\right)
+
\sum_{m=1}^{M}
\exp\left(\operatorname{sim}(q_i,k^-_m)/\tau\right)
}.
\end{multline}
where $\tau$ is the temperature parameter and $\operatorname{sim}(\cdot,\cdot)$ denotes cosine similarity.

For the ground-truth class $y$, we define the target-class query and negative sets as
\begin{equation}
Q_y = \{u_i \mid i \in \mathcal{P}_y\}, \qquad
\mathcal{M}_y = \{u_j \mid j \in \mathcal{N}_y\},
\end{equation}
and formulate the target-class loss as
\begin{equation}
\mathcal{L}_y = \mathcal{L}_{\mathrm{InfoNCE}}(Q_y, t_y, \mathcal{M}_y).
\end{equation}

To regularize the semantic structure beyond the target class, for each non-target class $c \neq y$, we define
\begin{equation}
Q_c = \{u_i \mid i \in \mathcal{P}_c\}, \qquad
\mathcal{M}_c = \{u_j \mid j \in \mathcal{N}_c\},
\end{equation}
and compute the symmetric auxiliary loss
\begin{equation}
\mathcal{L}_{sym}
=
\frac{1}{C-1}
\sum_{c \neq y}
\mathcal{L}_{\mathrm{InfoNCE}}(Q_c, t_c, \mathcal{M}_c).
\end{equation}

To further enhance inter-class discrimination, for each non-target class $c \neq y$, we compute an auxiliary prototype from the mean representation of the bottom-ranked projected features:
\begin{equation}
p_{y \rightarrow c}
=
\frac{1}{|\mathcal{N}_c|}
\sum_{j \in \mathcal{N}_c} u_j.
\end{equation}
The cross-class separation loss is then defined as
\begin{equation}
\mathcal{L}_{cross}
=
\frac{1}{C-1}
\sum_{c \neq y}
\mathcal{L}_{\mathrm{InfoNCE}}(Q_y, p_{y \rightarrow c}, Q_c),
\end{equation}
which encourages target-relevant projected features to remain close to representations that are dissimilar to non-target classes while staying separated from projected features that are highly relevant to those competing classes.

The overall concept-guided representation learning objective is
\begin{equation}
\mathcal{L}_{\mathrm{CGRL}}
=
\mathcal{L}_{y}
+
\lambda_{sym}\mathcal{L}_{sym}
+
\lambda_{cross}\mathcal{L}_{cross}.
\end{equation}
Let $\mathcal{L}_{cls}$ denote the bag-level classification loss produced by the downstream MIL model. The final training objective is
\begin{equation}
\mathcal{L}
=
\mathcal{L}_{cls}
+
\lambda_{\mathrm{CGRL}}\mathcal{L}_{\mathrm{CGRL}}.
\end{equation}
During inference, the contrastive objective is not used, and the model performs only concept-relevance pruning followed by MIL-based slide classification.

\begin{table*}[t]
\centering
\caption{Quantitative results on TCGA-BRCA and NSCLC datasets. Within each method group, the best results are shown in \textbf{bold} and the second-best are \underline{underlined}.}
\label{tab:combined_results}
\resizebox{\textwidth}{!}{%
\begin{tabular}{l|ccc|ccc}
\toprule
\multirow{2}{*}{\textbf{Method}} 
& \multicolumn{3}{c|}{\textbf{TCGA-BRCA}} 
& \multicolumn{3}{c}{\textbf{NSCLC}} \\
\cmidrule(lr){2-4} \cmidrule(lr){5-7}
& \textbf{AUC} & \textbf{ACC} & \textbf{F1}
& \textbf{AUC} & \textbf{ACC} & \textbf{F1} \\
\midrule

\cellcolor{gray!20}\textbf{ABMIL (Full)} 
&\cellcolor{gray!20} 0.9564 $\pm$ 0.0270 &\cellcolor{gray!20} 0.9101 $\pm$ 0.0061 &\cellcolor{gray!20} 0.8449 $\pm$ 0.0212
&\cellcolor{gray!20} \underline{0.9857 $\pm$ 0.0118} &\cellcolor{gray!20} \underline{0.9377 $\pm$ 0.0279} &\cellcolor{gray!20} \underline{0.9370 $\pm$ 0.0280} \\

ABMIL + Random 
& \textbf{0.9686 $\pm$ 0.0295} & \underline{0.9284 $\pm$ 0.0196} & \underline{0.8744 $\pm$ 0.0412}
& \textbf{0.9870 $\pm$ 0.0087} & 0.9280 $\pm$ 0.0252 & 0.9271 $\pm$ 0.0253 \\

ABMIL + Kmeans 
& 0.9576 $\pm$ 0.0243 & 0.9216 $\pm$ 0.0220 & 0.8562 $\pm$ 0.0425
& 0.9848 $\pm$ 0.0102 & \textbf{0.9437 $\pm$ 0.0242} & \textbf{0.9429 $\pm$ 0.0242} \\

ABMIL + CGRL (ours) 
& \underline{0.9660 $\pm$ 0.0286} & \textbf{0.9307 $\pm$ 0.0156} & \textbf{0.8781 $\pm$ 0.0254}
& 0.9731 $\pm$ 0.0236 & 0.9282 $\pm$ 0.0336 & 0.9271 $\pm$ 0.0341 \\

\midrule

\cellcolor{gray!20}\textbf{CLAM\_SB (Full)} 
&\cellcolor{gray!20} \textbf{0.9674 $\pm$ 0.0183} & \cellcolor{gray!20}0.9306 $\pm$ 0.0174 &\cellcolor{gray!20} 0.8733 $\pm$ 0.0428
&\cellcolor{gray!20} \textbf{0.9882 $\pm$ 0.0099} &\cellcolor{gray!20} \textbf{0.9529 $\pm$ 0.0229} &\cellcolor{gray!20} \textbf{0.9523 $\pm$ 0.0230} \\

CLAM\_SB + Random 
& 0.9607 $\pm$ 0.0236 & 0.9283 $\pm$ 0.0163 & 0.8704 $\pm$ 0.0473
& \underline{0.9864 $\pm$ 0.0118} & 0.9280 $\pm$ 0.0258 & 0.9270 $\pm$ 0.0261 \\

CLAM\_SB + Kmeans 
& \underline{0.9649 $\pm$ 0.0277} & \underline{0.9332 $\pm$ 0.0290} & \underline{0.8748 $\pm$ 0.0682}
& 0.9841 $\pm$ 0.0128 & \underline{0.9356 $\pm$ 0.0336} & \underline{0.9349 $\pm$ 0.0339} \\

CLAM\_SB + CGRL (ours) 
& 0.9591 $\pm$ 0.0245 & \textbf{0.9400 $\pm$ 0.0102} & \textbf{0.8982 $\pm$ 0.0159}
& 0.9731 $\pm$ 0.0427 & 0.9205 $\pm$ 0.0756 & 0.9197 $\pm$ 0.0756 \\

\midrule

\cellcolor{gray!20}\textbf{TransMIL (Full)} 
&\cellcolor{gray!20} 0.9352 $\pm$ 0.0280 &\cellcolor{gray!20} \underline{0.9150 $\pm$ 0.0328} &\cellcolor{gray!20} \underline{0.8592 $\pm$ 0.0471}
&\cellcolor{gray!20} 0.9781 $\pm$ 0.0078 &\cellcolor{gray!20} 0.9273 $\pm$ 0.0325 &\cellcolor{gray!20} 0.9263 $\pm$ 0.0323 \\

TransMIL + Random 
& \underline{0.9581 $\pm$ 0.0279} & 0.9147 $\pm$ 0.0212 & 0.8560 $\pm$ 0.0331
& \underline{0.9825 $\pm$ 0.0085} & \underline{0.9313 $\pm$ 0.0227} & \underline{0.9304 $\pm$ 0.0223} \\

TransMIL + Kmeans 
& \textbf{0.9582 $\pm$ 0.0341} & \textbf{0.9214 $\pm$ 0.0217} & \textbf{0.8602 $\pm$ 0.0482}
& \textbf{0.9854 $\pm$ 0.0145} & 0.9305 $\pm$ 0.0214 & 0.9287 $\pm$ 0.0235 \\

TransMIL + CGRL (ours) 
& 0.9510 $\pm$ 0.0314 & 0.9122 $\pm$ 0.0191 & 0.8433 $\pm$ 0.0337
& 0.9771 $\pm$ 0.0271 & \textbf{0.9322 $\pm$ 0.0408} & \textbf{0.9315 $\pm$ 0.0408} \\

\midrule

\cellcolor{gray!20}\textbf{DS\_MIL (Full)} 
&\cellcolor{gray!20} 0.9358 $\pm$ 0.0540 &\cellcolor{gray!20} 0.9120 $\pm$ 0.0412 &\cellcolor{gray!20} \underline{0.8490 $\pm$ 0.0693}
&\cellcolor{gray!20} 0.9780 $\pm$ 0.0200 &\cellcolor{gray!20} 0.9266 $\pm$ 0.0324 &\cellcolor{gray!20} 0.9251 $\pm$ 0.0330 \\

DS\_MIL + Random 
& \underline{0.9467 $\pm$ 0.0324} & 0.9122 $\pm$ 0.0172 & 0.8429 $\pm$ 0.0487
& \underline{0.9862 $\pm$ 0.0089} & \underline{0.9319 $\pm$ 0.0297} & \underline{0.9310 $\pm$ 0.0296} \\

DS\_MIL + Kmeans 
& \textbf{0.9551 $\pm$ 0.0290} & \textbf{0.9259 $\pm$ 0.0303} & \textbf{0.8687 $\pm$ 0.0607}
& \textbf{0.9895 $\pm$ 0.0106} & \underline{0.9418 $\pm$ 0.0172} & \underline{0.9411 $\pm$ 0.0174} \\

DS\_MIL + CGRL (ours) 
& 0.9414 $\pm$ 0.0349 & \underline{0.9125 $\pm$ 0.0236} & 0.8439 $\pm$ 0.0379
& 0.9836 $\pm$ 0.0144 & \textbf{0.9490 $\pm$ 0.0223} & \textbf{0.9486 $\pm$ 0.0221} \\

\bottomrule
\end{tabular}
}
\end{table*}

\begin{table}[!t]
\centering
\caption{Ablation study of different loss components on the TCGA-BRCA dataset. 
The best results are shown in \textbf{bold} and the second-best are \underline{underlined}.}
\label{tab:ablation_loss}
\resizebox{\columnwidth}{!}{
\begin{tabular}{lccc}
\toprule
\textbf{Method} & \textbf{AUC} & \textbf{ACC} & \textbf{F1} \\
\midrule
\cellcolor{gray!20}Baseline (ABMIL) 
&\cellcolor{gray!20} 0.9564 $\pm$ 0.0270 
&\cellcolor{gray!20} 0.9101 $\pm$ 0.0061 
&\cellcolor{gray!20} 0.8449 $\pm$ 0.0212 \\

w/ $L_y$ 
& 0.9530 $\pm$ 0.0419 
& 0.9074 $\pm$ 0.0377 
& 0.8164 $\pm$ 0.1000 \\

w/ $L_{sym}$ 
& 0.9533 $\pm$ 0.0332 
& \underline{0.9258 $\pm$ 0.0293} 
& 0.8644 $\pm$ 0.0533 \\

w/ $L_{cross}$ 
& 0.9588 $\pm$ 0.0330 
& 0.9165 $\pm$ 0.0319 
& 0.8641 $\pm$ 0.0481 \\

w/ $L_y + L_{sym}$ 
& 0.9564 $\pm$ 0.0358 
& 0.9210 $\pm$ 0.0393 
& 0.8553 $\pm$ 0.0815 \\

w/ $L_y + L_{cross}$ 
& \textbf{0.9738 $\pm$ 0.0194} 
& 0.9238 $\pm$ 0.0121 
& \underline{0.8720 $\pm$ 0.0226} \\

w/ $L_{sym} + L_{cross}$ 
& 0.9602 $\pm$ 0.0186 
& 0.9124 $\pm$ 0.0292 
& 0.8566 $\pm$ 0.0351 \\

\midrule
Baseline + CGRL 
& \underline{0.9660 $\pm$ 0.0286} 
& \textbf{0.9307 $\pm$ 0.0156} 
& \textbf{0.8781 $\pm$ 0.0254} \\
\bottomrule
\end{tabular}
}
\end{table}

% \begin{table}[!t]
% \centering
% \caption{Ablation study of different loss components on the TCGA-BRCA dataset. 
% The best results are shown in \textbf{bold} and the second-best are \underline{underlined}.}
% \label{tab:ablation_loss}
% \resizebox{\columnwidth}{!}{
% \begin{tabular}{lcccc}
% \toprule
% \textbf{Method} & \textbf{AUC} & \textbf{ACC} & \textbf{F1} & \textbf{Kappa} \\
% \midrule
% \cellcolor{gray!20}Baseline (ABMIL) 
% &\cellcolor{gray!20} 0.9564 $\pm$ 0.0270 
% &\cellcolor{gray!20} 0.9101 $\pm$ 0.0061 
% &\cellcolor{gray!20} 0.8449 $\pm$ 0.0212 
% &\cellcolor{gray!20} 0.6905 $\pm$ 0.0434 \\

% w/ $L_y$ 
% & 0.9530 $\pm$ 0.0419 
% & 0.9074 $\pm$ 0.0377 
% & 0.8164 $\pm$ 0.1000 
% & 0.6360 $\pm$ 0.1948 \\

% w/ $L_{sym}$ 
% & 0.9533 $\pm$ 0.0332 
% & \underline{0.9258 $\pm$ 0.0293} 
% & 0.8644 $\pm$ 0.0533 
% & 0.7289 $\pm$ 0.1065 \\

% w/ $L_{cross}$ 
% & 0.9588 $\pm$ 0.0330 
% & 0.9165 $\pm$ 0.0319 
% & 0.8641 $\pm$ 0.0481 
% & 0.7296 $\pm$ 0.0953 \\

% w/ $L_y + L_{sym}$ 
% & 0.9564 $\pm$ 0.0358 
% & 0.9210 $\pm$ 0.0393 
% & 0.8553 $\pm$ 0.0815 
% & 0.7112 $\pm$ 0.1624 \\

% w/ $L_y + L_{cross}$ 
% & \textbf{0.9738 $\pm$ 0.0194} 
% & 0.9238 $\pm$ 0.0121 
% & \underline{0.8720 $\pm$ 0.0226} 
% & \underline{0.7447 $\pm$ 0.0456} \\

% w/ $L_{sym} + L_{cross}$ 
% & 0.9602 $\pm$ 0.0186 
% & 0.9124 $\pm$ 0.0292 
% & 0.8566 $\pm$ 0.0351 
% & 0.7149 $\pm$ 0.0694 \\

% \midrule
% Baseline + CGRL 
% & \underline{0.9660 $\pm$ 0.0286} 
% & \textbf{0.9307 $\pm$ 0.0156} 
% & \textbf{0.8781 $\pm$ 0.0254} 
% & \textbf{0.7568 $\pm$ 0.0510} \\
% \bottomrule
% \end{tabular}
% }
% \end{table}

%% file: sections/05_experiments.tex
\section{Experiments}

\subsection{Datasets}
We conduct experiments on two TCGA datasets obtained from the NCI Genomic Data Commons (GDC): TCGA-BRCA and TCGA-NSCLC \cite{heath2021gdc}. 
The latter includes lung adenocarcinoma (LUAD) and lung squamous cell carcinoma (LUSC) cohorts. 
TCGA-BRCA contains 871 whole-slide images (WSIs) from two breast cancer subtypes, including 723 invasive ductal carcinoma (IDC) samples and 148 invasive lobular carcinoma (ILC) samples. 
TCGA-NSCLC contains 958 WSIs from two lung cancer subtypes, including 492 LUAD samples and 466 LUSC samples. 
For both datasets, we adopt 5-fold cross-validation for evaluation.

\subsection{MIL-based Baselines} 
% \textcolor{red}{phan nay em viet mo ta ABMIL, cLAM-SB, TransMIL va DS-MIL}

We compare CGRL with four representative MIL baselines for weakly supervised WSI classification: ABMIL, CLAM-SB, TransMIL, and DS-MIL. These methods represent different bag modeling strategies, including attention-based pooling, clustering-constrained learning, Transformer-based correlation modeling, and dual-stream aggregation. \cite{ilse2018attention,lu2021data,li2021dual,shao2021transmil}

\textbf{ABMIL}\cite{ilse2018attention}: uses a trainable attention-based pooling mechanism to aggregate patch embeddings into a slide-level representation. It learns an attention weight for each instance and computes the bag representation as a weighted sum of instance features, providing both effective aggregation and interpretable patch importance. 

\textbf{CLAM-SB}\cite{lu2021data}: extends ABMIL by adding an instance-level clustering constraint. It uses highly attended and weakly attended patches as representative positive and negative evidence to improve the discriminative structure of the feature space. 

\textbf{TransMIL}\cite{shao2021transmil}: models correlations among patch instances using a Transformer-based architecture. By combining self-attention with positional encoding, it captures both morphological and spatial dependencies across the whole slide. 

\textbf{DS-MIL} \cite{li2021dual}: adopts a dual-stream design in which one stream identifies the critical instance with the highest score, while the other aggregates all instances according to their similarity to this critical instance. This design is effective for weakly supervised WSI classification, especially in imbalanced bags.

\subsection{Implementation Details}
For preprocessing, each whole-slide image is first subjected to foreground tissue segmentation and then divided into $256 \times 256$ patches at $10\times$ magnification. 
We use the pretrained \textbf{CONCH v1.5} image encoder~\cite{lu2024conch} to extract a 768-dimensional feature representation for each patch. 
To construct the class-level concept prototypes, class-specific disease prompts are encoded by the pretrained \textbf{TITAN} text encoder~\cite{ding2025multimodal}, yielding 768-dimensional concept embeddings. 
The prompt construction strategy and the full prompt set follow \textbf{CATE}~\cite{huang2024free}. 
Both pretrained encoders are kept frozen throughout training.

For concept-relevance pruning, we retain the top-$K_p$ most relevant patches with $K_p=300$ for all experiments.

The model is optimized using Adam with a learning rate of $2\times10^{-4}$ and a batch size of 1. 
For the loss configuration, we set $\lambda_{\mathrm{CGRL}}=0.01$, $\lambda_{sym}=1$, and $\lambda_{cross}=1$. 
For contrastive sample construction, we set $\mathrm{topk}_{pos}=\mathrm{topk}_{neg}=20$. 
All experiments are conducted on a single NVIDIA GeForce RTX 2080 Ti GPU.

%% file: sections/06_discussion.tex
\section{Results}
\subsection{Main Results}

Table~\ref{tab:combined_results} summarizes the quantitative results on TCGA-BRCA and TCGA-NSCLC. 
Overall, the results reveal two important observations. 
First, patch sampling is often beneficial: in several settings, the sampled variants outperform the corresponding full-bag baselines, suggesting that removing redundant or weakly informative patches can improve both robustness and slide-level discrimination. 
This observation is consistent with the motivation of our method, since WSIs typically contain a large number of non-informative patches and pruning them can help the downstream MIL model focus on more relevant evidence.

Second, among the sampling-based variants, CGRL delivers strong and often superior performance in several important model--dataset combinations, especially on threshold-dependent metrics such as ACC and macro-F1. 
On TCGA-BRCA, CGRL is particularly effective for attention-based MIL models. 
For example, ABMIL + CGRL improves ACC/F1 from 0.9101/0.8449 to 0.9307/0.8781, while CLAM-SB + CGRL achieves the best ACC/F1 within the CLAM family, reaching 0.9400/0.8982. 
On TCGA-NSCLC, the effect of concept-guided sampling is more model-dependent, but CGRL still provides clear gains for more challenging models. 
In particular, TransMIL + CGRL improves ACC/F1 from 0.9273/0.9263 to 0.9322/0.9315, and DS-MIL + CGRL yields the most notable improvement, increasing ACC/F1 from 0.9266/0.9251 to 0.9490/0.9486.

At the same time, the results also indicate that the effectiveness of sampling is model- and dataset-dependent.
In some cases, Random or Kmeans sampling remains competitive, and for several AUC results the full model or alternative sampling strategy may still perform better. 
Nevertheless, CGRL shows the clearest advantages on class-discriminative metrics in multiple representative settings, suggesting that concept-guided patch selection provides a more semantically informative subset for downstream MIL aggregation.

\subsection{Ablation study}

As shown in Table~\ref{tab:ablation_loss}, the results indicate that combining $L_y$ and $L_{cross}$ significantly improves AUC, 
highlighting the importance of cross-class separation. 
Meanwhile, the full CGRL framework achieves the best overall performance in ACC and F1, 
demonstrating the complementary effects of all loss components. 
These results suggest that $L_y$ and $L_{cross}$ contribute strongly to class separation, whereas integrating all objectives leads to a more balanced and discriminative representation for slide-level classification.

%% file: sections/07_conclusion.tex
\section{Conclusion}
In this paper, we proposed \emph{Concept-Guided Pruning and Representation Learning} (CGRL), a simple and effective framework for weakly supervised whole-slide image classification. 
The proposed method introduces class-level concept prototypes to guide both concept-relevance pruning and representation learning, enabling the model to identify more concept-relevant instances before slide-level aggregation. 
By coupling concept-relevance pruning with concept-guided contrastive supervision, CGRL provides a unified way to select a semantically more relevant subset of patches for downstream MIL while regularizing projected concept-space representations during training.Since concept-guided contrastive supervision is used only during training, inference retains only concept-relevance pruning and MIL-based slide classification.

Extensive experiments on TCGA-BRCA and TCGA-NSCLC demonstrate that CGRL can serve as a practical plug-and-play module for multiple MIL methods. 
In particular, CGRL achieves clear improvements in several method--dataset combinations, with especially notable gains in accuracy and macro-F1, indicating better slide-level discrimination and stronger class-wise agreement. 
At the same time, the results also show that the effectiveness of concept guidance is method-dependent, suggesting that the interaction between semantic priors and bag aggregation strategies plays an important role in the final performance. 
Overall, these findings highlight the potential of leveraging textual medical concepts to enhance weakly supervised pathology learning.

In future work, we plan to explore more expressive concept construction strategies, stronger vision--language alignment, and adaptive concept selection mechanisms to further improve generalization across datasets and MIL architectures.